\documentclass[10pt,twocolumn,letterpaper]{article}

\usepackage{iccv}
\usepackage{times}
\usepackage{epsfig}
\usepackage{graphicx}
\usepackage{amsmath}
\usepackage{amssymb}
\usepackage{caption}
\usepackage{multirow}
\usepackage{algorithmic}
\usepackage[vlined,ruled]{algorithm2e} 
\SetKwProg{Fn}{Function}{}{}

\usepackage[breaklinks=true,bookmarks=false]{hyperref}

\iccvfinalcopy % *** Uncomment this line for the final submission

\def\iccvPaperID{****} % *** Enter the ICCV Paper ID here

\ificcvfinal\fi
\begin{document}

%%%%%%%%% TITLE
% \title{GAT: Gated Multi-Level Attention with Temporal Adversarial Training for Video Understanding}
\title{Enhancing Transformer for Video Understanding Using Gated Multi-Level Attention and Temporal Adversarial Training}
\author{Saurabh Sahu\\
Samsung Research America\\
{\tt\small saurabh.s1@samsung.com}
\and
Palash Goyal\\
Samsung Research America\\
{\tt\small palash.goyal@samsung.com}
}

\makeatletter
\let\@oldmaketitle\@maketitle
\renewcommand{\@maketitle}{\@oldmaketitle

    \begin{tabular}{cc}
  \includegraphics[width=0.49\textwidth]{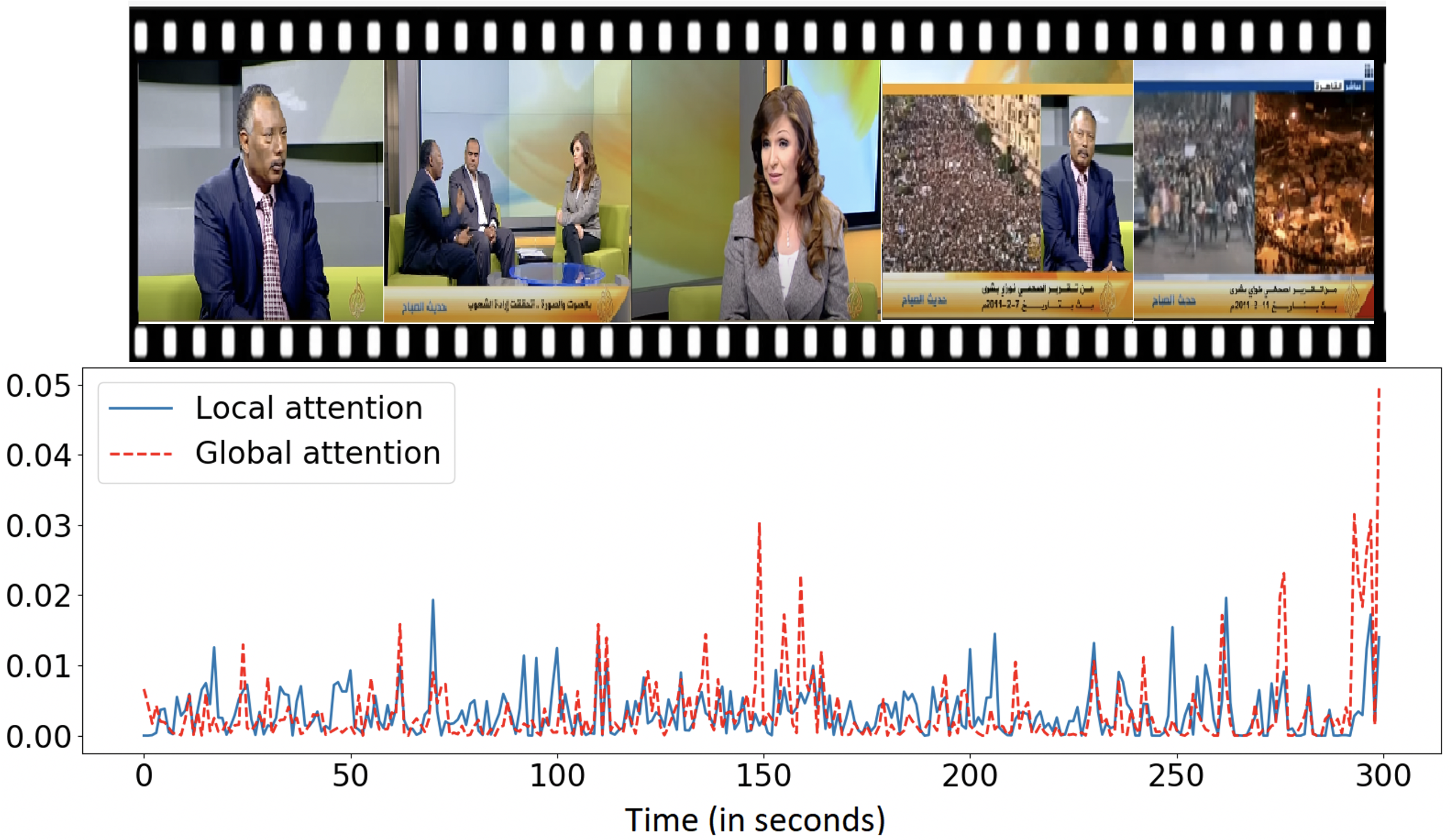}
  &
  \includegraphics[width=0.49\textwidth]{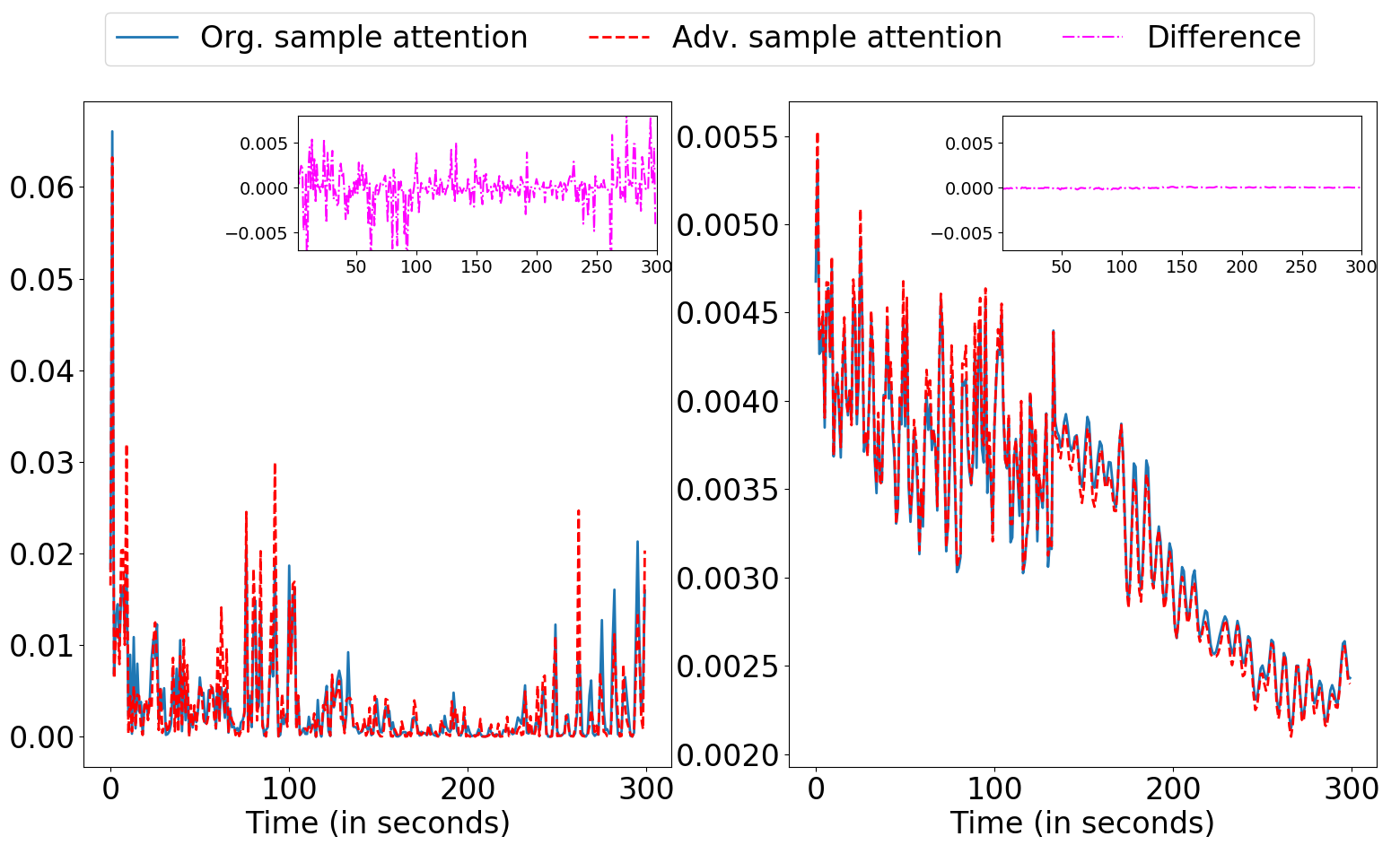}
  \\
  (a) Attention Profiles for a Video & (b) Attention Profiles for Original and Adversarial Video 
    \vspace{-0.2cm}
  \end{tabular}
      \captionof{figure}{\textbf{Gated Adversarial Transformer Advantages.} (a) Local and global attention profiles obtained for a `News' video in YouTube-8M (youtube.com/watch?v=ygORXiV2Zpw). Global attention profile is more uneven and puts most attention towards the end of the video giving an incorrect prediction (`Sports') while the local attention profile is more uniform temporally. Using multi-level gated attention, we get the correct prediction (`News'). (b) Attention map generated by the models trained without adversarial learning (left) and our proposed model (right) for a test sample and its adversarial counterpart. Our model enhances robustness to adversarial examples in attention as well as output space.}
   %\caption{Figure 1. \textbf{Gated Adversarial Transformer Advantages.} (a) Local and global attention profiles obtained for a `News' video in YouTube-8M (youtube.com/watch?v=ygORXiV2Zpw). Global attention profile is more uneven and puts most attention towards the end of the video giving an incorrect prediction (`Sports') while the local attention profile is more uniform temporally. Using multi-level gated attention, we get the correct prediction (`News'). (b) Attention map generated by the models trained without adversarial learning (left) and our proposed model (right) for a test sample and its adversarial counterpart. Our model enhances robustness to adversarial examples in attention as well as output space.}
  \label{fig:gat}
  \bigskip}
\makeatother
\maketitle
\ificcvfinal\thispagestyle{empty}\fi
\begin{abstract}
\vspace{-5mm}
The introduction of Transformer model has led to tremendous advancements in sequence modeling, especially in text domain. However, the use of attention-based models for video understanding is still relatively unexplored. In this paper, we introduce Gated Adversarial Transformer (GAT) to enhance the applicability of attention-based models to videos. GAT uses a multi-level attention gate to model the relevance of a frame based on local and global contexts. This enables the model to understand the video at various granularities. Further, GAT uses adversarial training to improve model generalization. We propose temporal attention regularization scheme to improve the robustness of attention modules to adversarial examples. We illustrate the performance of GAT on the large-scale YoutTube-8M data set on the task of video categorization. We further show ablation studies along with quantitative and qualitative analysis to showcase the improvement.
\end{abstract}

\section{Introduction}
\label{sec:intro}
Video classification aims at understanding the visual and audio features to assign one or more relevant tags to the video. With the rapid increase in amount of video content, this task is crucial for applications such as smart content search, user profiling and alleviating missing metadata. Moving from image to video classification adds several challenges to the task. First, the temporal dimension increases the overall size of the input in turn increasing the model capacity required to make accurate predictions. Secondly, the number of possible tags increase due to variations in sequence. For e.g., a leaf falling from tree can be tagged as nature but the reverse could indicate science-fiction elements in the video.

Existing works on video classification using deep learning can be broadly divided into four categories: (i) convolutional neural networks (CNNs)~\cite{tran2019video, lin2019tsm, carreira2017quo, tran2015learning}, (ii) recurrent neural networks (RNNs)~\cite{zhu2020faster, yang2017tensor, wu2017deep}, (iii) graph-based methods~\cite{mao2018hierarchical, aminianvidsage}, and (iv) attention-based models~\cite{sahu2020cross,kmiec2018learnable, girdhar2019video}. Convolutional approaches use 3-D CNN along with optical flow networks to jointly model spatial and temporal features. Recurrent models use variations of recurrent neural cell including gated recurrent unit (GRU) and long-short term memory (LSTM) to understand the video. Graph based approaches create a similarity graph of frames to cluster scenes together. Lastly, attention-based models use self-attention blocks at each layer to understand a frame's role with respect to other frames in the video. In this paper, we focus on attention approaches as they have low computation cost compared to 3-D CNNs and outperform RNN based approaches on video classification task~\cite{sahu2020cross}.

Current attention-based approaches suffer from two key issues which we aim to tackle in this paper. First, video representation and classification is highly dependent on the attention weights. This can lead to incomplete predictions if high attention is given to only a handful of relevant frames, and incorrect predictions if given to irrelevant frame(s). For e.g., in Figure 1 (left), for a news video on protests, we show attention weights and prediction of Transformer model trained on 4 million YouTube videos. We observe that high attention is given to the frames containing crowd and thus final prediction is given as sports (confusing it with sports crowd). Second issue which has recently been showed for videos~\cite{wei2019sparse} is that the models are prone to adversarial examples. In Figure 1 (right), we observe that a minor change in input modifies attention profiles obtained from a self-attention based architecture.%In Figure 1 (right), we show the attention profiles of normal and adversarial examples and observe that a minor change in input modifies the output of encoder.

In this paper, to alleviate the above issues, we propose Gated Adversarial Transformer (GAT) which makes two technical contributions. (1) a multi-level gated attention: instead of using a single global profile we define a local and global attention profile and use a gated mixture of experts model to learn a weighted combination of the representations obtained from the two profiles. (2) temporal adversarial training: we compute the adversarial direction for each example and add an attention-regularization term to ensure robustness to adversarial example at two levels: attention level and output classification.

We make the following contributions in this paper:
\begin{itemize}
    \item We observed global attention dependency problem in state-of-the-art self-attention based models. We propose a gated \textit{multi-level attention module} to tackle this.
    \item We propose \textit{temporal adversarial training} for video understanding to enhance the robustness of attention-based encoders.
    \item We provide comprehensive \textit{large-scale quantitative and qualitative experiments} on YouTube-8M data, showing significant improvements with our approach over state-of-the-art modes. 
\end{itemize}
\section{Related Work}
\textbf{Attention based models}:
With the success of attention based architectures (such as Transformers) in NLP tasks, recently they have been used in video related tasks as well. The self-attention block proposed in~\cite{vaswani2017attention} is closely related to the non-local blocks used for video classification~\cite{wang2018non}. Chen et al.~\cite{chen20182} proposed increasing the efficacy of existing convolutional blocks for action classification by using a two-stage attention mechanism to collect relevant features from different patches of video frames at various instants. Kmiec et al.~\cite{kmiec2018learnable} used Transformer encoder along with NetVLAD blocks to get effective feature representations of audio/video features for large scale video understanding. Wu et al.~\cite{wu2019long} used an attention block to compute the interactions between short-term and long-term feature representations for detailed video understanding. Girdhar et al.~\cite{girdhar2019video} proposed `action Transformers' where features extracted using a 3D CNN are aggregated and fed to a self-attention block to leverage the spatio-temporal information around a person for action localization. Bertasius et al.~\cite{bertasius2021space} extended the idea of~\cite{dosovitskiy2020image} to apply Transformers for video classification by collectively using the patches obtained from a frame at different time-steps. While the prior works have used vanilla self-attention block's operations, we propose modifications to self-attention architecture itself to make it more relevant for video understanding. Furthermore, it can also be used in cases where raw videos are not available for training (such as Youtube-8M dataset) thereby rendering some of the above methods infeasible as they rely on extracting information from image patches.

\textbf{Adversarial training}:
It has been shown that machine learning models misclassify correctly classified images when a small non-perceptible perturbation is added to them ~\cite{szegedy2013intriguing, 43405, moosavi2016deepfool}. ~\cite{wei2019sparse} showed it for videos where the trained model fails to detect the correct class of a perturbed video. For a threat model, they generated the adversarial perturbations in an iterative way by maximising the cross-entropy loss between the model's output for a perturbed video and its ground-truth label. They additionally minimize the $L_{21}$ norm of the perturbations so that the perturbed video is semantically close to the original video. This ensured that even though both the videos are visually similar, the model outputs differ from each other. In ~\cite{43405, miyato2018virtual}, the authors propose a way of computing the adversarial counterparts of images while training and adding an extra loss regularization that forces the model to correctly classify them or make their predictions close to that of the original image. We extend this idea to videos. As we are using a self-attention based architecture, we additionally implement a attention regularization term in the loss function that promotes similarity in the attention map between original and adversarial videos, ensuring robustness in both attention and output space.
\section{Gated Adversarial Transformer (GAT)}
\begin{figure}[!hbtp]
\centering
    \includegraphics[width=0.7\columnwidth]{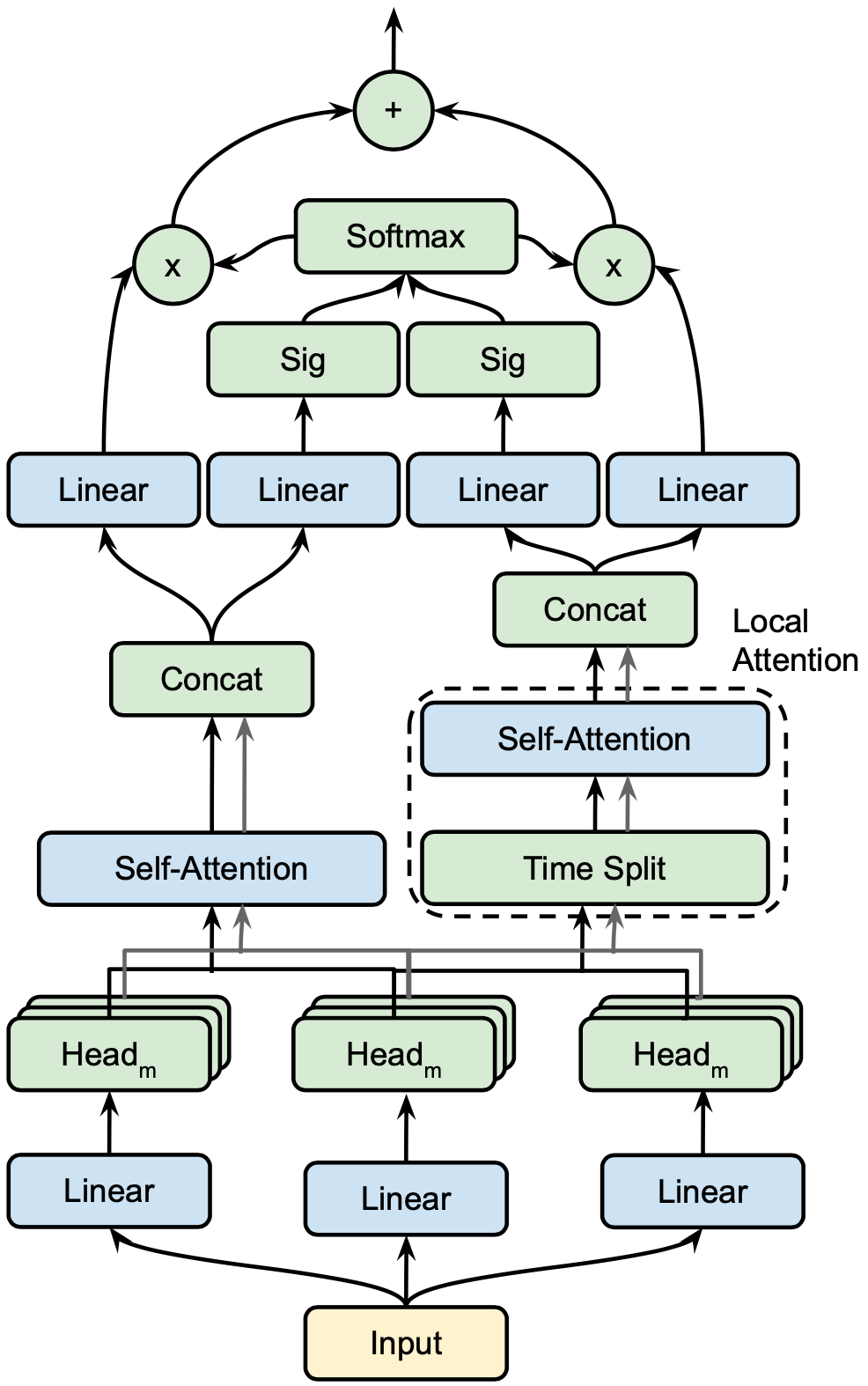}
    \caption{Gated Multi-level Self Attention (GMSA) architecture.}
    \label{fig:gated_arch}
    \vspace{-0.4cm}
\end{figure}
In this section, we explain our model architecture and training algorithm. Given $D$ dimensional input features for $T$ frames of a video, we encode the temporal information with a positional encoding block to get a representation $X \in \mathbb{R}^{T\times D}$ and pass it through a multi-head self attention (SA) block to get the output representations. This is followed by layer-normalization, residual connections and feed-forward layers as proposed in ~\cite{vaswani2017attention}. We then get a video level representation by taking a mean across the temporal dimension. We get these video level representations for both the audio and visual modalities. Then they are concatenated and passed through a hidden layer before getting the final output. We incorporate novelties in the way we get the output representation from the SA block. Additionally, we train the model using regularization loss terms computed from adversarial examples along with the vanilla cross-entropy loss term.
\subsection{Gated Multi-level Self Attention}
For a video, the inter-frame relationship is different when seen in local context versus global context. Hence, to identify a video correctly, we compute multiple output representations from local as well as global attention maps. The final output representation is computed by taking a weighted combination of these representations, the weights being determined by a soft-gating mechanism. Figure~\ref{fig:gated_arch} outlines the gated multi-level attention architecture. 

In a multi-headed SA block with $M$ heads, input $X$ is first transformed into query, key and value matrices. For $m$-th head, we compute the attention map $A_m \in \mathbb{R}^{T\times T}$ using the scaled dot product of the corresponding query $Q_m \in \mathbb{R}^{T\times D_M}$ and key $K_m \in \mathbb{R}^{T\times D_M}$ matrices. It is then multiplied with the value matrix $V_m \in \mathbb{R}^{T\times D_M}$ to get a global level output representation for the head $O_{gm} \in \mathbb{R}^{T\times D_M}$. We call it a global level representation because it is compute using the attention map that depicts the importance of a frame $i$ with respect to frame $j$ taking into context the entire video. These outputs from different heads heads are then concatenated and a linear transformation is done to get the final output representation $Y^g \in \mathbb{R}^{T\times D}$. 
\begin{eqnarray}
Q_m &=& XW^q_m, \hspace{1mm} K_m = XW^k_m, \hspace{1mm}V_m = XW^v_m \nonumber\\
A_m &=& \mbox{softmax} \big(\frac{Q_mK_m^T}{\sqrt{d_m}}\big)\label{eqn:2}\\%\nonumber\\
O_{gm} &=& A_m.V_m \label{transformer}\nonumber\\
Y^g &=& \mbox{concat}(O_{g1},\ldots,O_{gM}).W^g\nonumber 
\nonumber
\end{eqnarray}
\noindent where $W^q_m \in \mathbb{R}^{D\times D_M}$, $W^k_m \in \mathbb{R}^{D\times D_M}$, $W^v_m \in \mathbb{R}^{D\times D_M}$ and $W^g \in \mathbb{R}^{D\times D}$ are learnable matrices. Note that in this formulation, only a global attention map is computed by each head to get the final output representation. 

We additionally compute local attention maps and use it to influence the final output representation of the block. Specifically, the query, key and value matrices are divided into $N$ segments each of duration $T_N = T/N$. For the $n$-th segment in $m$-th head the query matrix $Q_{n,m} \in \mathbb{R}^{T_N\times D_M} $ is obtained by taking the frames $\{n*T_N,\ldots,(n+1)*T_N-1\}$ of the matrix $Q_m$. Similarly we get the localised key $K_{n,m} \in \mathbb{R}^{T_N\times D_M}$ and value $V_{n,m} \in \mathbb{R}^{T_N\times D_M}$ matrices. Using these matrices, we get the output $O_{n,m} \in \mathbb{R}^{T_N\times D_M} $ as shown in the equation below. These are then concatenated in the temporal dimension to get the local level representation for the head $O_{lm} \in \mathbb{R}^{T\times D_M} $ followed by getting the final local level representation $Y^l \in \mathbb{R}^{T\times D}$ by multiplying with a learnable matrix $W^l \in \mathbb{R}^{D\times D}$
\begin{eqnarray}
O_{n,m} &=& \mbox{softmax} \big(\frac{Q_{n,m}K_{n,m}^T}{\sqrt{d_m}}\big).V_{n,m} \label{gated_transformer} \nonumber \\
O_{lm} &=& \mbox{concat}(O_{1,m},\ldots,O_{N,m})\nonumber\\
Y^l &=& \mbox{concat}(O_{l1},\ldots,O_{lM}).W^l \nonumber 
\nonumber
\end{eqnarray}
This local level output representation can carry relevant information about the video. It is used along with the global level representation to get the final output. We incorporate a gating mechanism that considers the multi-level output representations as different experts and computes the final output representation which is fed to subsequent layers for video classification. We call it gated multi-level self-attention (GMSA). Specifically the relevance $R^g \in \mathbb{R}^{T\times D}$ and $R^l \in \mathbb{R}^{T\times D}$ are computed using the gating network. This is followed by computing the final output $Y \in \mathbb{R}^{T\times D}$ as shown below
\begin{eqnarray}
R^g &=& \mbox{concat}(O_{g1},\ldots,O_{gM}).W^g_g \nonumber \\
R^l &=& \mbox{concat}(O_{l1},\ldots,O_{lM}).W^l_g \nonumber \\
R^g, R^l &=& softmax([R^g,R^l]) \nonumber \\
Y &=& R^g \odot Y^g + R^l \odot Y^l \nonumber 
\nonumber
\end{eqnarray}
\noindent where $W^g_g \in \mathbb{R}^{D\times D}$, $W^g_l \in \mathbb{R}^{D\times D}$ are learnable layers and $\odot$ denotes element-wise multiplication. In the gating network, softmax is taken across the experts similar to ~\cite{shazeer2017outrageously}. In the above multi-level formulation we have considered two output representations -- one at local and one at global level. Note that we can get multiple such attention representations for different values of $N$ each of them acting as an expert to get the relevance weights for them.  

Since, for local level representations, the softmax is computed over a time period of $T_N$ instead of $T$, it gives us additional information about how a frame is perceived in relation to a local context than the entire global context. This extra information is complimentary to the global level context which is shown in our experiments in section~\ref{sec:exp}. 
\subsection{Adversarial Perturbation based Regularization}
Having changed the architecture of the self-attention block to take into account attentions computed at different granularity, we now focus on modifying the loss function for more robust learning. In ~\cite{wei2019sparse}, the authors showed adding a small amount of perturbation to an existing video which doesn't change the semantics of video, can fool a pre-trained video classification model. We plan to address this vulnerability of deep learning based video classification models. Below, we formalize our approach.

We denote the training set with $L$ data point as $\{X_i, y_i\}, i = 1,..,L $, where $X_i \in \mathbb{R}^{T\times D}$ represents the frame-wise feature representation of video $i$ and $y_i \in \mathbb{R}^{K}$ represents the ground-truth labels. $K$ is the number of classes. We represent the video classification model's output vector of probabilities for the point $X_i$ as $\theta (X_i)$.
%Hence, $\theta (X_i)$ is a vector of probabilities that the model assigns to each class in the label space spanned by $y$.
$l(\theta(X_i), y_i)$ is the loss for the data point $X_i$ which we consider to be cross-entropy loss in our multi-class scenario. 
\begin{equation}
\mathcal L_{CE} = \frac{1}{L} \sum_{i=1}^L l(y_i, \theta(X_i)) \label{eqn:ce}
\end{equation}
However, such models would be susceptible to adversarial examples hurting the generalizability of the model. To address this limitation, we add a regularization term minimizing the loss for adversarial counterparts of the training samples as proposed in ~\cite{43405}. 
\begin{eqnarray}
\mathcal L_{Adv} &=& \mathcal L_{CE} + \alpha * \frac{1}{L} \sum_{i=1}^L l(y_i, \theta(X_i+R_i)) \nonumber \\
R_i &=& \arg\max_{R_i:\|R_i\|_2\le\epsilon}l(y_i, \theta(X_i+R_i)) \nonumber \\
\nonumber
\end{eqnarray}
The loss function is approximated to behave linearly around the input $X_i$ to get the perturbation term which can be easily calculated using backpropagation. 
\begin{equation}
R_i \approx \epsilon \frac{G_i}{\|G_i\|_2}, where\    G_i = \nabla_{X_i}l(y_i, \theta(X_i)) \label{eqn:per}
\end{equation}

In the above equations, the norm is taken across rows of the input tensor. We compute the gradient of the loss with respect to the features from each of video and audio modalities to get the corresponding adversarial features.
Note that, we train the model to be invariant to adversarial
samples within the $\epsilon$ ball. Hence, optimizing this loss function has two hyper-parameters to tune, $\alpha$ and $\epsilon$. We investigate the impact of these hyper-parameters on the model performance in our experiments.
\subsubsection{Attention-map regularization}
The attention-map being generated in the self-attention blocks should also be invariant to adversarial examples. We add a regularization term in our loss function to enforce this condition. For a given input, we average over the attention maps $A_m$ generated by each head to get the attention map $A \in \mathbb{R}^{T\times T}$ for that input. The corresponding attention map generated using the adversarial example is denoted by $A^{adv}$. Based on how the similarity is enforced in the attention space, we propose two variations of the adversarial loss function:
\begin{itemize}
\item We minimize the Frobenius norm of the difference between the attention maps and average over the mini-batch in which case the loss function becomes
\begin{equation}
\mathcal L_{advFr} = \mathcal L_{adv} + \beta_{Fr} * \frac{1}{L} \sum_{i=1}^L \|A_i - A^{adv}_i\|_{Fr} \nonumber \\
\end{equation}

\item Each row of the attention map can be treated as a probability distribution. Since Jensen-Shannon (JS) divergence between two distributions is symmetric, we minimize it to enforce similarity between the attention maps.
\begin{equation}
\mathcal L_{advJS} = \mathcal L_{adv} + \beta_{JS} * \frac{1}{LT} \sum_{i=1}^L \sum_{t=1}^T JSD(A_{i,t}, A^{adv}_{i,t}) \nonumber \\
\end{equation}
where $A_{i,t}$ and $A^{adv}_{i,t}$ denote the $t$-th row of the attention-map generated by the $i$-th example in the min-batch and its adversarial counterpart respectively.
\end{itemize}
%As we get two sets of attention maps corresponding to the two modalities, we take their average to get the final attention map regularization term in the loss functions mentioned above. 
Note that we can use both local and global attention maps computed by the GMSA block. However, we observed that there was not a significant change in performance when using both the attention maps for regularization. Hence, we only show results for the global attention map regularization in our experiment section.

We call the model combining gated multi-level self attention (GMSA) and adversarial perturbation with attention regularization as Gated Adversarial Transformer (GAT).
\subsubsection{Algorithm}
The overall algorithm describing our training procedure given features from audio and video modalities and the ground truth labels is described below. For a mini-batch, we compute the cross-entropy loss and adversarial perturbations using equations~\ref{eqn:ce} and ~\ref{eqn:per} respectively. Then, the function $get\_att$ gets the global attention map averaged over the heads computed using Equation~\ref{eqn:2}. We compute the loss for adversarial samples and the attention-map regularization term and get the final loss $L$ to update the model parameters.
\begin{algorithm}[!htb]\color{black}
\Fn{gatTraining}{
Initialise $\theta^{enc}_v$, $\theta^{enc}_a$ and $\theta_{MLP}$ \\
\For{$e=1 \ldots E$}{
%$L_{CE} = 0$, $L_{CE}^{adv} = 0$, $L_{Fr} = 0$\\
\For{$b=1 \ldots B$}{
$\hat{y_b} = \theta_{MLP}(\theta^{enc}_v(X_{b,v}), \theta^{enc}_a(X_{b,a}))$ \\
$L_{CE} = l(y_b, \hat{y_b}) $ \\
\For{$m$ in $\{a,v\}$}{
$R_{b,m} = \epsilon \frac{\nabla_{X_{b,m}}L_{CE}}{\|\nabla_{X_{b,m}}L_{CE}\|_2}$\\
%$R_{b,m}$ = stop\_gradient($R_{b,m}$)\\
$X_{b,m}^{ad}$ = $X_{b,m}$ + $R_{b,m}$\\
$A_{b,m} = get\_att(\theta^{enc}_m, X_{b,m})$\\
$A_{b,m}^{ad} = get\_att(\theta^{enc}_m, X_{b,m}^{ad})$\\
}
$L_F = \|A_{b,v} - A^{ad}_{b,v}\|+\|A_{b,a} - A^{ad}_{b,a}\|$\\
$\hat{y_b^{ad}} = \theta_{MLP}(\theta^{enc}_v(X_{b,v}^{ad}), \theta^{enc}_a(X_{b,a}^{ad}))$ \\
%$L_{CE} = L_{CE} + L_b$\\
$L_{CE}^{ad} = l(y_b, \hat{y_b^{ad}}) $\\
$L = L_{CE} + \alpha*L_{CE}^{ad} + \beta_{Fr}*L_F$\\
$\theta^{enc}_v = \theta^{enc}_v - \eta*\nabla_{\theta^{enc}_v}L$ \\
$\theta^{enc}_a = \theta^{enc}_a - \eta*\nabla_{\theta^{enc}_a}L$ \\
$\theta_{MLP} = \theta_{MLP} - \eta*\nabla_{\theta_{MLP}}L$ \\
}
}
return $\theta^{enc}_v$, $\theta^{enc}_a$ and $\theta_{MLP}$}
\caption{Training GAT with Frobenius attention map regularization given input audio-visual features $(X_v, X_a)$, ground truth labels $y$, loss function $l$, gated multi-level attention based Transformer encoder blocks $\theta^{enc}_v$ and $\theta^{enc}_a$ for the two modalities, MLP parameters $\theta_{MLP}$ , some radius $\epsilon$, hyper-parameters $\alpha$ and $\beta_{Fr}$, learning rate $\eta$ for $E$ epochs with mini-batch size $B$}
\label{alg:s_selection}
\end{algorithm}

\section{Experiments}
\label{sec:exp}
\begin{table*}[!htbp]
\centering
\caption{Overall performance (in percentages) of baselines and different variations of our proposed model for video categorization task on our YouTube-8M test sets. Each column represents the results for a training paradigm defined by the architecture and the loss function used. We present the mean and standard deviation obtained for five non-overlapping partitions of the entire test set.}
\label{tab:yt8m_res}
% \scriptsize
\vspace{-3mm}
\begin{tabular}{c||cccccc}
\hline
 Metrics   &  SA, $\mathcal L_{CE}$   & GMSA, $\mathcal L_{CE}$ & GMSA, $\mathcal L_{Adv}$& GAT$_{AdvJS}$ & GAT$_{AdvFr}$& Gain\\ \hline
GAP       &    91.48$\pm$0.02 &  92.18$\pm$0.02 & 92.49$\pm$0.02& 92.56$\pm$0.02 & 92.60$\pm$0.02  &\textbf{+1.12}              \\ 
MAP       & 91.29$\pm$0.02   & 92.32$\pm$0.03 & 92.68$\pm$0.01 & 92.70$\pm$0.04 & 92.80$\pm$0.03 & \textbf{+1.51}       \\ 
PERR     &   89.46$\pm$0.04 & 90.03$\pm$0.04& 90.34$\pm$0.04& 90.41$\pm$0.03 & 90.44$\pm$0.03 & \textbf{+0.98}           \\
Hit@1    &  94.84$\pm$0.05  & 95.02$\pm$0.04& 95.16$\pm$0.03 & 95.21$\pm$0.05 & 95.22$\pm$0.04& \textbf{+0.38}         \\\hline
\end{tabular}
\vspace{-3mm}
\end{table*}
In this section, we explain our experiments and results. We first show our experimental setup. Then, we present the results for video classification on a large-scale YouTube dataset. We further analyse the various novel aspects of our model and present quantitative and qualitative analysis.
\subsection{Experimental-setup}
We use YouTube-8M dataset for our experiments which consists of frame-wise video and audio features for approximately 5 million videos extracted using Inception v3 and VGGish respectively followed by PCA ~\cite{abu2016youtube}. We use the hierarchical label space with 431 classes (see \cite{sahu2020cross}). We use binary cross-entropy loss to train our models. We evaluate our models using the four metrics mentioned in ~\cite{lee20182nd}: (i) Global Average Precision (GAP), (ii) Mean Average Precision (MAP), (iii) Precision at Equal Recall Rate (PERR), and (iv) Hit@1. 
Our training set consists of approximately 4 million videos. We use 64000 videos from the official development set for validation and use the rest as test set. Our baseline Transformer model consists of a single layer of multi-head attention with 8 attention heads for each of audio and video modalities. For training we used Adam optimizer, with an initial learning rate of 0.0002 and batch size of 64. We compute validation set GAP every 10000 iterations and perform early-stopping with patience of 5.
We also use it for learning-rate scheduler that decreases the learning rate by a factor of 0.1 with patience of 3.
\subsection{Results}
We present the results for video categorization using a baseline Transformer encoder with and without our novelties. Specifically, we compare the performance of SA with GMSA and using  cross-entropy loss function $\mathcal L_{CE}$.  The window size $T_N$ for the GMSA module was set to be $20$ based on the best GAP on validation set. We choose the best performing architecture and implement the adversarial loss function $\mathcal L_{Adv}$.  Based on validation set results, the value of $\epsilon$ and $\alpha$ for adversarial training was set to be $0.5$ and $1$ respectively. $\beta_{Fr}$ and $\beta_{JS}$ are set to $0.001$ and $0.01$ respectively. To test the variance in performance of the different models under different test set distributions, we divide our test set into five equal partitions. We compute the mean and standard deviation of each of the four metrics obtained across these five partitions.

From Table~\ref{tab:yt8m_res}, we see that using GMSA block in Transformer outperforms SA block in all metrics. Improvement in MAP is more compared to GAP and Hit$@$1 suggesting that the performance boost is more for underrepresented classes. Moreover, using adversarial loss $\mathcal L_{adv}$, we note an improvement in performance compared to baseline cross entropy. Furthermore, enforcing attention map regularization between the original and adversarial examples leads to an even better performance showing the applicability of the extra regularization term. 
% Enforcing that the model parameters generate similar attention maps for both original and adversarial samples while training, leads to better optimization.
We obtain slightly better results with the Frobenius norm compared to JS divergence for attention regularization.

%We note that across the different test sets, there is more variance in Hit@1 compared to GAP and MAP. Hit@1 captures improvement in top 1 prediction while GAP and MAP capture improvement in top 20 making the former more prone to variations in test-set distributions.

\subsection{Analysis}
We first provide hyperparameter analysis and qualitative analysis for gated mult-level self attention. We then analyze the effect of attention map regularization on model robustness and performance. We further show qualitative analysis and class-wise analysis of the final model.
\subsubsection{Gated multi-level self attention}
We present the GAP values on validation set of varying local attention window length ($T_N$) in Figure~\ref{fig:gated}. We also show results if both the experts are given global attention map, which still improves over the baseline SA architecture since the two experts can learn complementary information. Further, we show results with three gating experts --- one global attention map and two local attention maps.
\renewcommand{\thefigure}{3}
\begin{figure}[hbtp]
\centering
    \includegraphics[width=0.9\columnwidth]{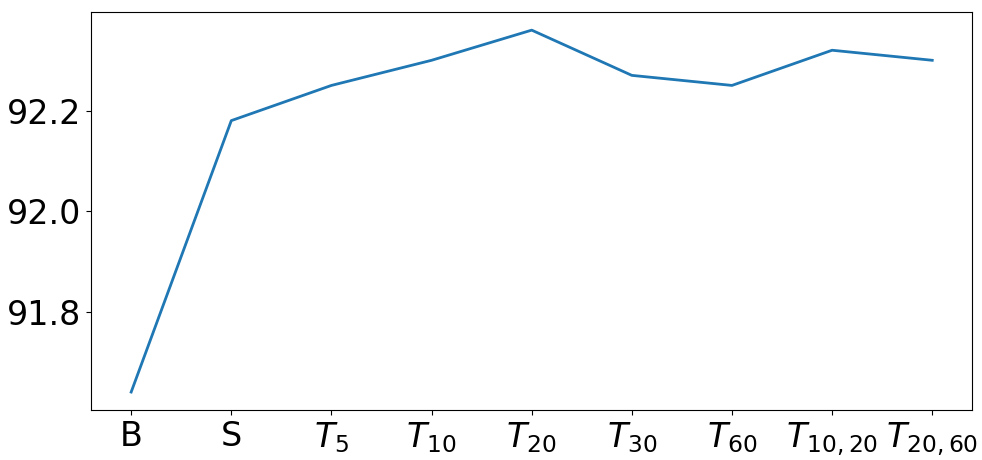}
    \caption{Validation GAP for (i) baseline SA architecture (B), (ii) GMSA with global attention as input to both experts (S), (iii) GMSA different values of window size $T_N$, and (iv) GMSA with three gating experts ($T_{N,M}$) --- global attention map and two local attention maps with window sizes $T_N$ and $T_M$.}
    \label{fig:gated}
    \vspace{-0.4cm}
\end{figure}

% \begin{figure}[hbtp]
% \centering
%     \includegraphics[width=1.0\columnwidth]{figures/global_2.png}
%     \caption{Local and global attention profiles obtained for a `News' video in test set (https://www.youtube.com/watch?v=ygORXiV2Zpw). Note that global attention profile is more uneven and puts most attention towards the end of the video giving an incorrect prediction (`Sports') while the local attention profile is more uniform temporally leading to a correct prediction (`News')}
%     \label{fig:gated_qual_2}
% \end{figure}
\renewcommand{\thefigure}{4}
\begin{figure}[hbtp]
\centering
    \includegraphics[width=1.0\columnwidth]{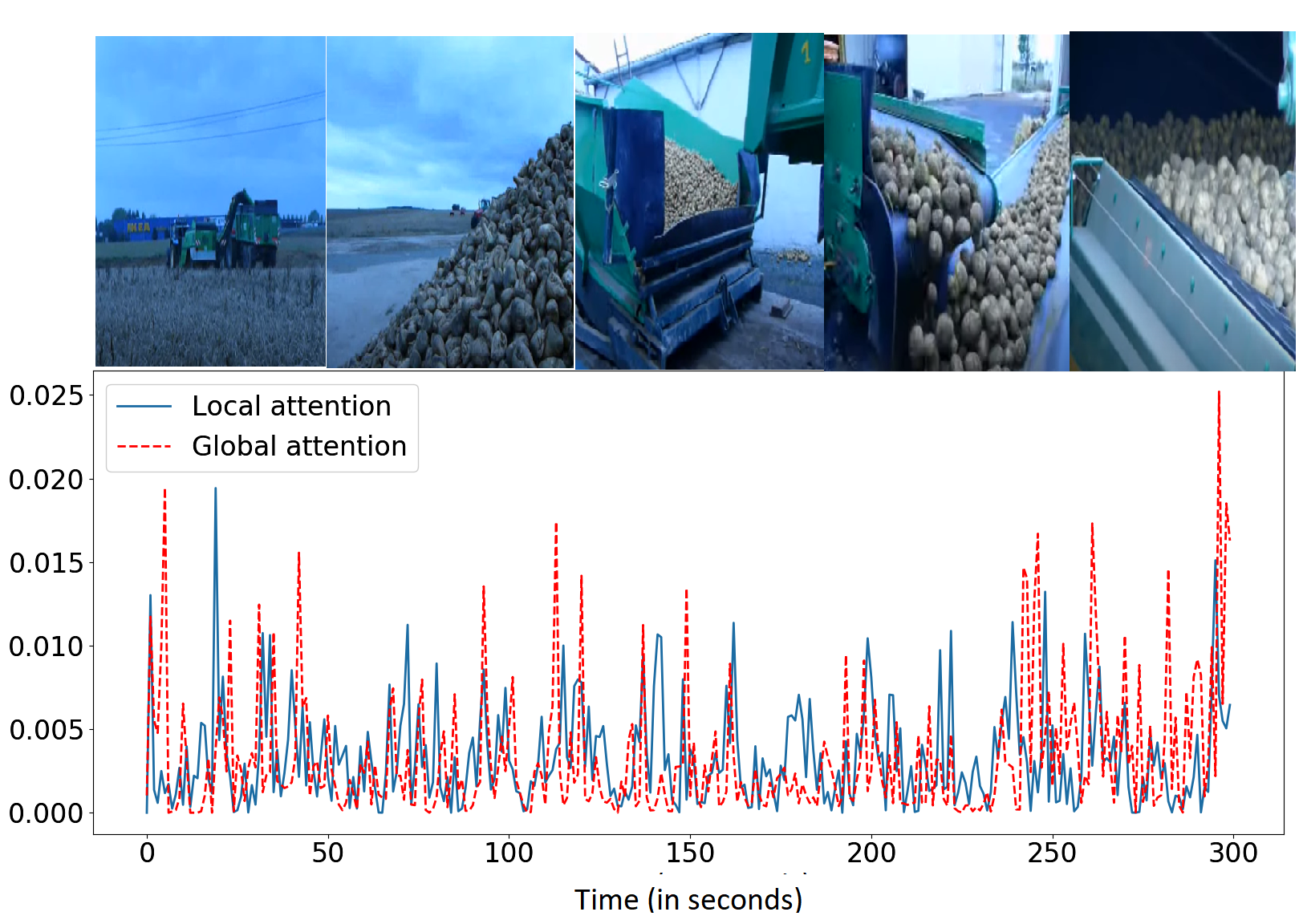}
    \caption{Local and global attention profiles obtained for a video in test set labeled as `Vegetables' (youtube.com/watch?v=uEXlc2kSXpM).}
    %While the baseline model can only identify broadly that the video is about `Food' with some probability, GMSA can accurately predict it is about `Vegetables' with high probability.
    \label{fig:gated_qual}
    \vspace{-0.5cm}
\end{figure}

We observe that using multiple gated experts improves the performance of a baseline SA block. Note that performance improves using both the local and global attention maps.
We also note that the validation set GAP saturates when more than two experts are used in GMSA.

We also compare the global attention profile obtained from a baseline SA block and the local attention profile from a GMSA block for two videos in Figure 1 and Figure~\ref{fig:gated_qual}. For global attention profile, we take mean over the rows of the 2D attention map to get an attention vector of length $T$. Similarly we get $N$ such local attention profiles each of length $T_N$ which are then concatenated to get the final local attention profile of length $T$. In Figure 1, we observe that the global attention profile peaks abruptly towards the end of the video which shows crowds chanting, possibly leading it to believe it to be a `Sports' video. The local attention profile attends the frames more evenly leading to the correct prediction that it is a `News' video. We observe that most of the frames in the video correspond to a studio setting with people having discussion. Hence, it makes sense that attending these frames uniformly leads to the correct prediction. 
In Figure ~\ref{fig:gated_qual}, while the baseline model can predict the presence of `Vehicle' with high probability ($0.7$) and that the video is about `Food' with some probability ($0.3$); it fails to predict the presence of `Vegetables' (probability $< 0.1$) which is actually the ground truth label of the video. The GMSA module can predict this video is about 'Vegetable' with a very high probability ($0.8$) along with `Vehicle' (with probability of $0.4$).
\subsubsection{Adversarial training}
\renewcommand{\thefigure}{5}
\begin{figure}[hbtp]
\centering
    \includegraphics[width=0.95\columnwidth]{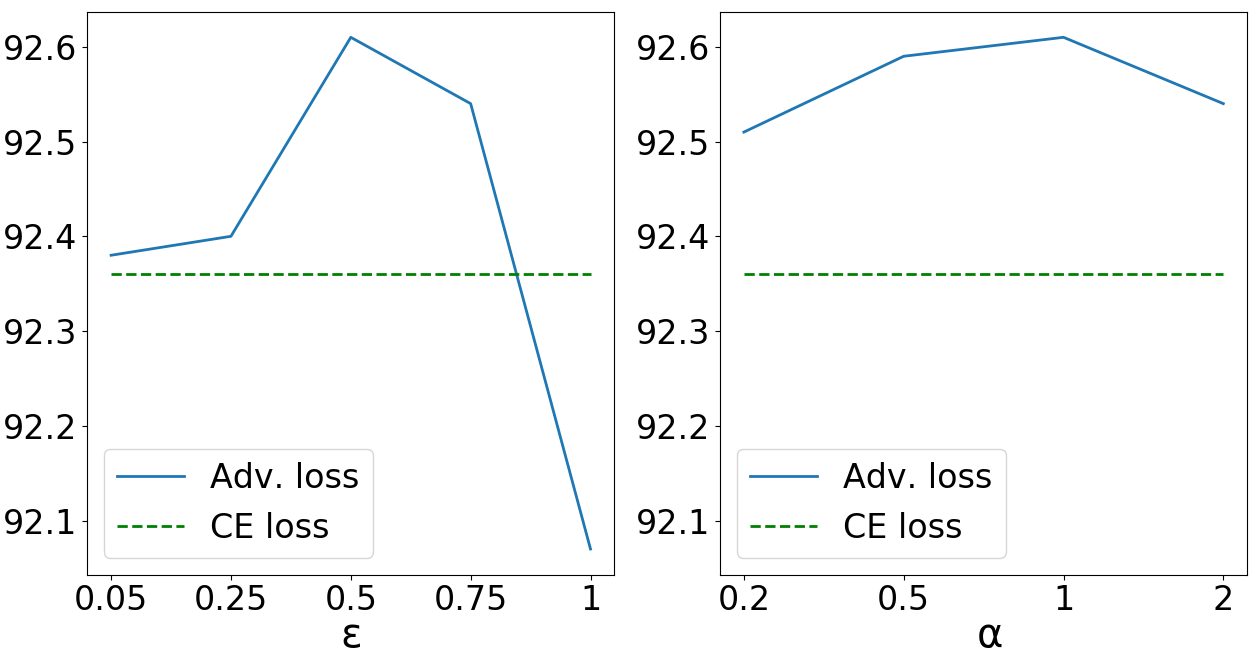}
    \caption{Validation GAP for models with cross-entropy loss and adversarial loss: neighborhood radius $\epsilon$ was varied with $\alpha = 1$(left) and $\alpha$ was varied with $\epsilon = 0.5$ (right).}
    \label{fig:AT}
    \vspace{-0.6cm}
\end{figure}
Based on experiments in the previous section, we use a GMSA based Transformer encoder block with $T_N = 20$ for our further experiments. We train the model with adversarial loss function $\mathcal L_{Adv}$. We aim to understand the impact of the two hyper-parameters $\epsilon$ and $\alpha$ on the model performance by perturbing one of them, while keeping the other constant. By altering $\epsilon$ , we aim to understand the impact of smoothing radius around the data-points on the model performance and perturbing $\alpha$ impacts the weight of the adversarial loss on the overall optimization. The plots comparing the validation GAP for different values of hyper-parameters is shown in Figure~\ref{fig:AT} . First, the value of $\alpha$ was kept fixed at 1 and $\epsilon$ was varied. For lower values of $\epsilon$, models trained with adversarial loss function $\mathcal L_{Adv}$ show an improvement over model trained with $\mathcal L_{CE}$ peaking at $\epsilon=0.5$.  As we increase the value of $\epsilon$, the model’s performance starts deteriorating. This is expected since $\epsilon$ defines the neighborhood around an input feature vector over which the conditional distribution is smoothed. Increasing the radius of this neighborhood forces our model to learn smoother functions that cannot capture the complexity of the conditional distribution function thereby decreasing its performance on the validation set. Similarly, as we increase the adversarial loss weight $\alpha$, the performance increases, peaks at $1$ and starts reducing as the relative weight of classification goes down.

%We note that training a model with the loss function $\mathcal L_{Adv}$ can be seen as an data augmentation method, where the augmented samples are obtained by adding adversarial perturbations to the original samples. To verify the importance of adversarial perturbations, we train a model with a loss function where the augmented samples were obtained by adding random perturbations than adversarial perturbations within the $\epsilon = 0.5$ ball. The validation GAP obtained with adding random perturbations was $92.3$ vs $92.6$ obtained with adversarial perturbation. This shows that adding perturbations in adversarial direction is crucial to improving model's performance.

We note that after using the attention map regularization, the Frobenius norm of the difference between the attention maps computed from the validation set samples and their adversarial counterparts reduces from $1.23$ to $0.02$. Qualitatively, the effect of using attention-map based regularisation in loss term ($\mathcal L_{advFr}$) can be seen in Figure~\ref{fig:ATvsATFr2}. Here, we compare the attention profiles being generated for a sample in test set and its adversarial counterpart by models trained using the vanilla adversarial loss $\mathcal L_{adv}$ and our approach $\mathcal L_{advFr}$. We observe that that attention generated by $\mathcal L_{advFr}$ is more robust to adversarial perturbations and the attention profiles of the original sample and its adversarial counterpart overlap to a great extent. On the other hand, the model trained with $\mathcal L_{adv}$ exhibits more variations in the attention profiles as a result of adversarial perturbations to the input. Another thing to notice from the figure is that the maximum attention being given to any frame reduces by an order of magnitude when using $\mathcal L_{advFr}$. In other words, the attention map generated by $\mathcal L_{advFr}$ is smoother enforcing the temporal coherence property which has been shown to help video classification model performance~\cite{huang2018makes, mobahi2009deep}.
\renewcommand{\thefigure}{6}
\begin{figure}[hbtp]
\centering
    \includegraphics[width=1.0\columnwidth]{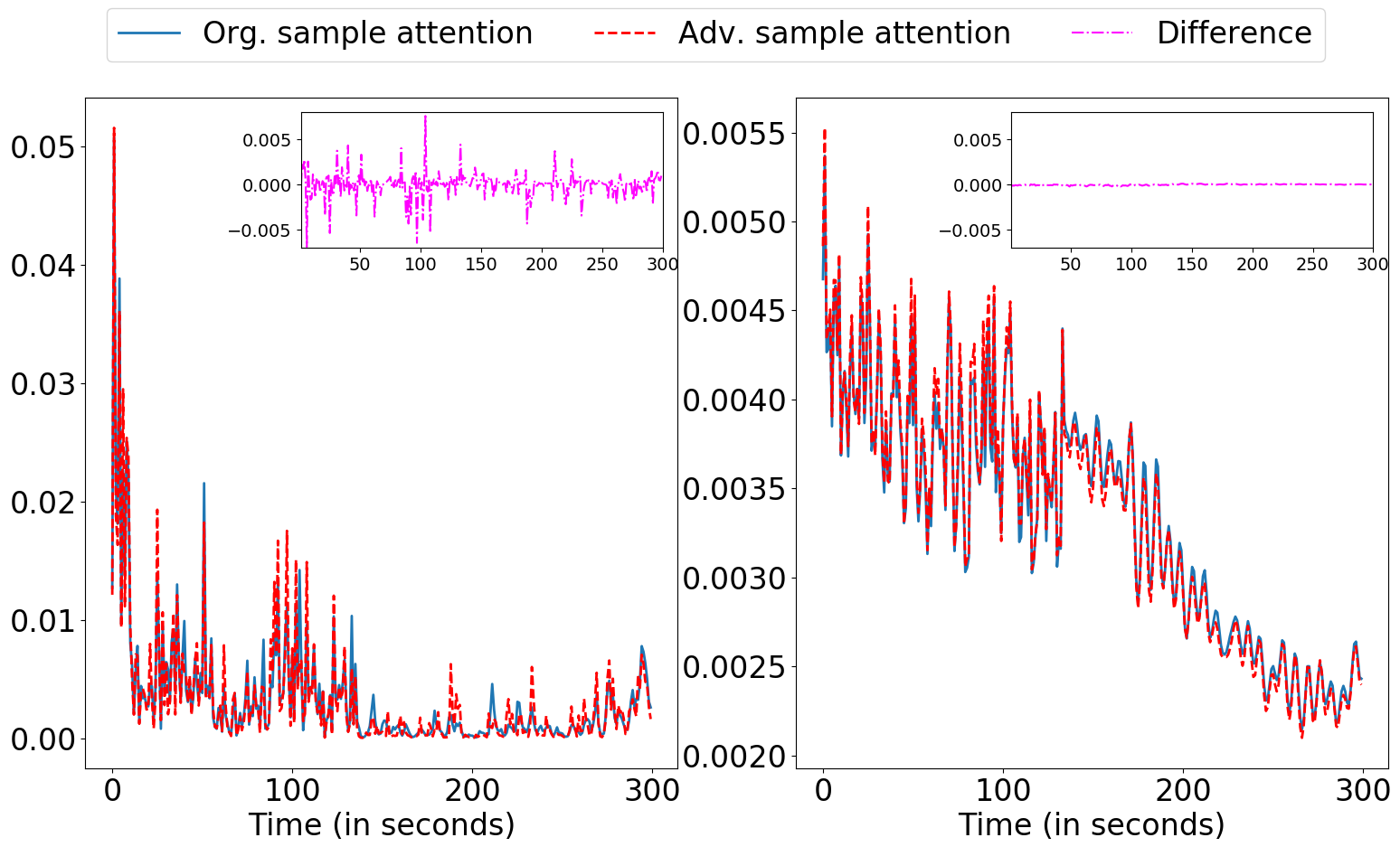}
    \caption{Attention profile generated by GMSA trained with $\mathcal L_{adv}$ (left) and $\mathcal L_{advFr}$ (right) for a test sample and its adversary.}
    \label{fig:ATvsATFr2}
    \vspace{-0.3cm}
\end{figure}
\begin{table*}[!ht]
\centering
\caption{Qualitative analysis for example videos in YouTube-8M comparing the performance of GAT with baseline Transformer model. Upto top 3 predictions with probabilities $> 0.2$ for each model are shown for comparison. Abbreviations MusicIns, StringIns, VG and Perf stand for MusicInstrument, StringInstrument, VideoGame and PerformanceArt respectively.}
\label{tab:qual}
% \scriptsize
\begin{tabular}{l|c|c|c|c}
            &    Example 1      &  Example 2      &   Example 3     &   Example 4 \\ \hline \hline
Video       &    \includegraphics[width=0.4\columnwidth]{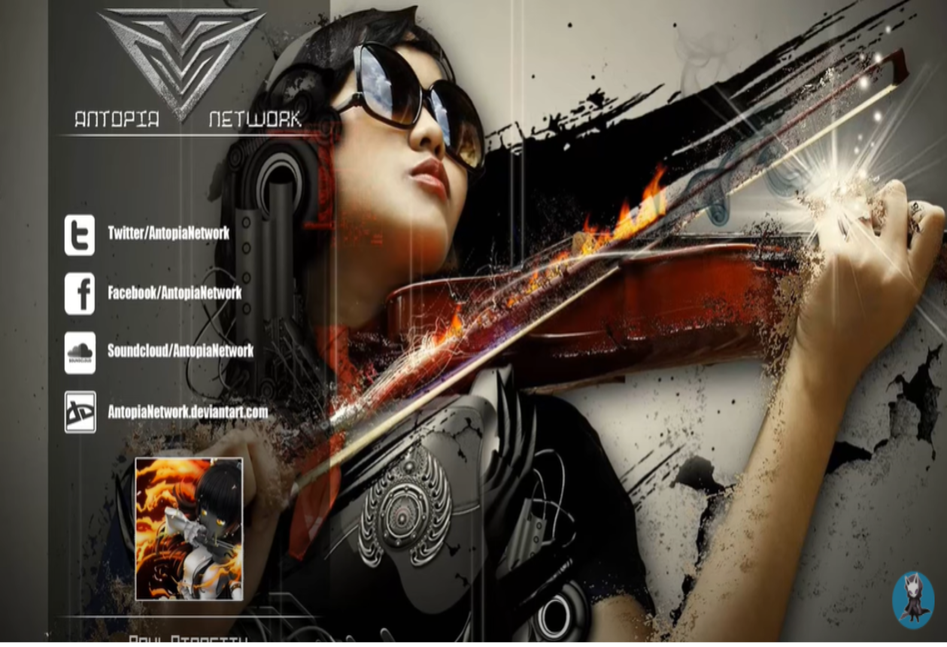}   &  \includegraphics[width=0.4\columnwidth]{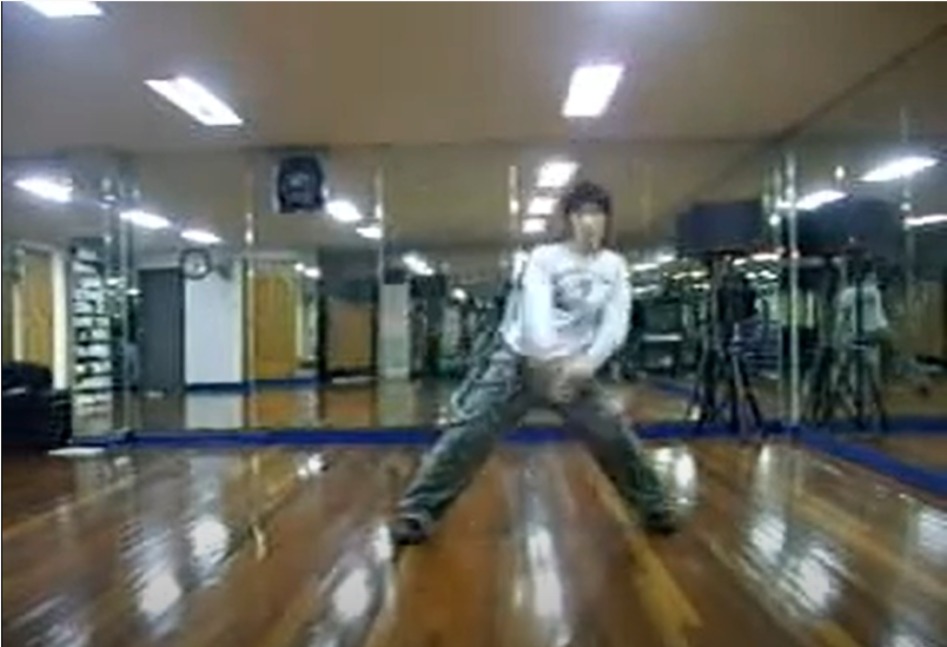}   &   \includegraphics[width=0.4\columnwidth]{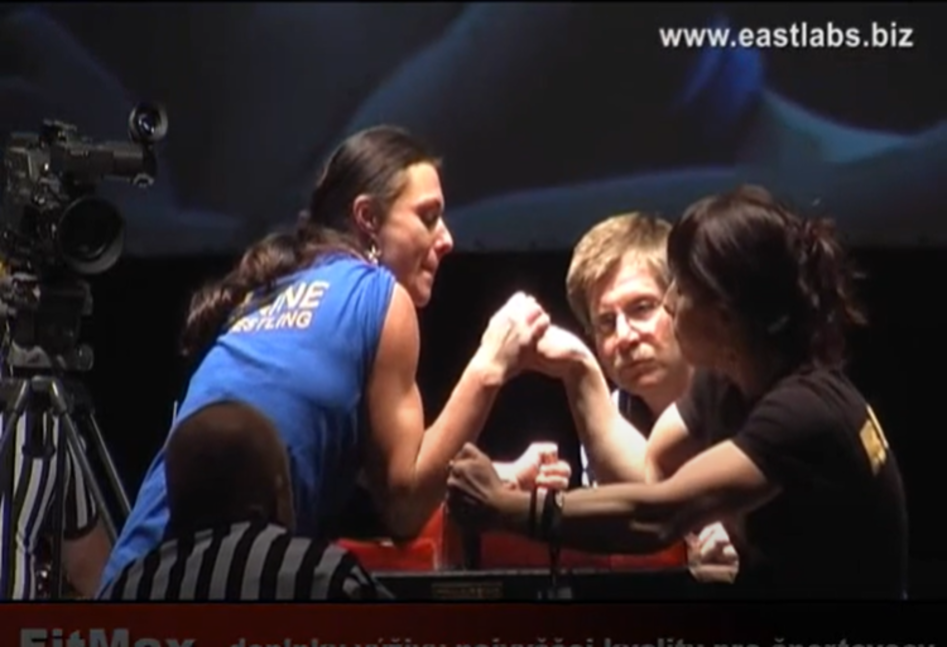}       &   \includegraphics[width=0.4\columnwidth]{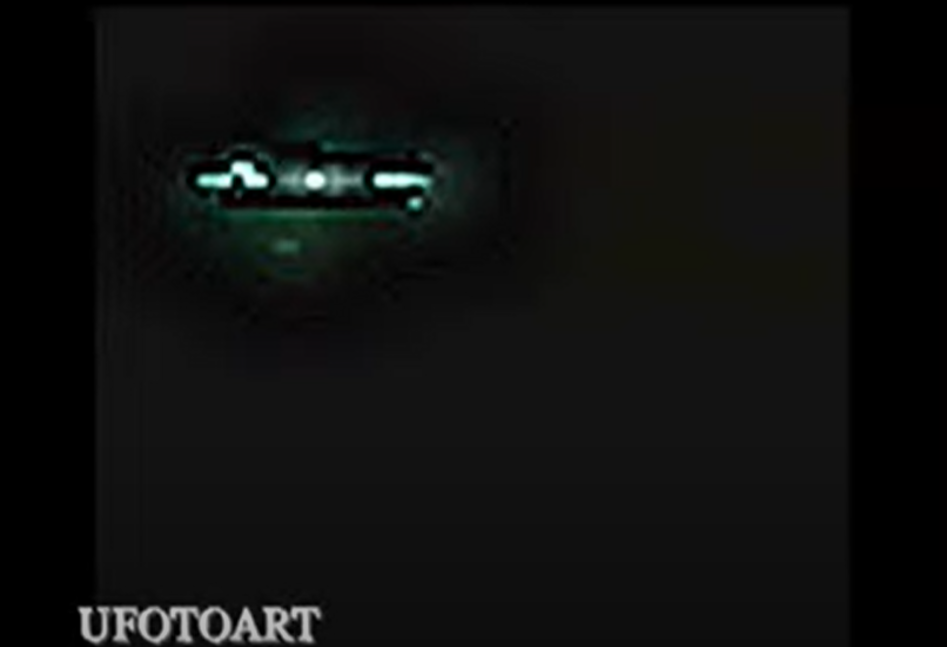}           \\ \hline\hline
Links & https://bit.ly/3bTxexj  & https://bit.ly/3lpTbr2 & https://bit.ly/3rWdMph & https://bit.ly/3qTOQNN \\ \hline \hline
      \multirow{2}{*}{GT} 
      		& Music &  &  & \\ 
      		& Music:MusicIns:StrIns & Art:Perf:Dance  & Sports:Combat:Wrestling  & Transport:Air:Spaceship \\ 
      	    \cline{1-5}\hline\hline
      \multirow{3}{*}{GAT} 
      		& &   & Sports:Combat: 0.65 & \\
      		& Music: 0.35& Art:Perf: 0.87  & Sports:Combat:Wrestling: 0.46 & Transport:Air:Spaceship: 0.91\\ 
      		& & Art:Perf:Dance: 0.78 & Music: 0.36  &  \\ \cline{1-5} \hline \hline
      \multirow{2}{*}{Baseline} 
      		& Game:VG: 0.77 & Sports: 0.83  &  & Game:VG: 0.7\\
      		& Game:VG:Action: 0.51 & Sports:Ball: 0.77 & Music: 0.6  & Game:VG:Action: 0.26 \\ 
      	  \cline{1-5}
  \hline
\end{tabular}
\vspace{-3mm}
\end{table*}

From Figure~\ref{fig:adv_attack}, we further notice that enforcing attention map regularization for adversarial samples makes the model more robust to adversarial attacks. For each of the saved models trained with losses $L_{CE}$, $\mathcal L_{adv}$ and $\mathcal L_{advFr}$, adversarial counterparts were created for the validation set using the current parameters. These were then used to evaluate the performance of the corresponding model. We notice that the models trained with $\mathcal L_{advFr}$ outperform those trained with $\mathcal L_{adv}$ and $\mathcal L_{CE}$ significantly.
\renewcommand{\thefigure}{7}
\begin{figure}[hbtp]
\centering
    \includegraphics[width=0.7\columnwidth]{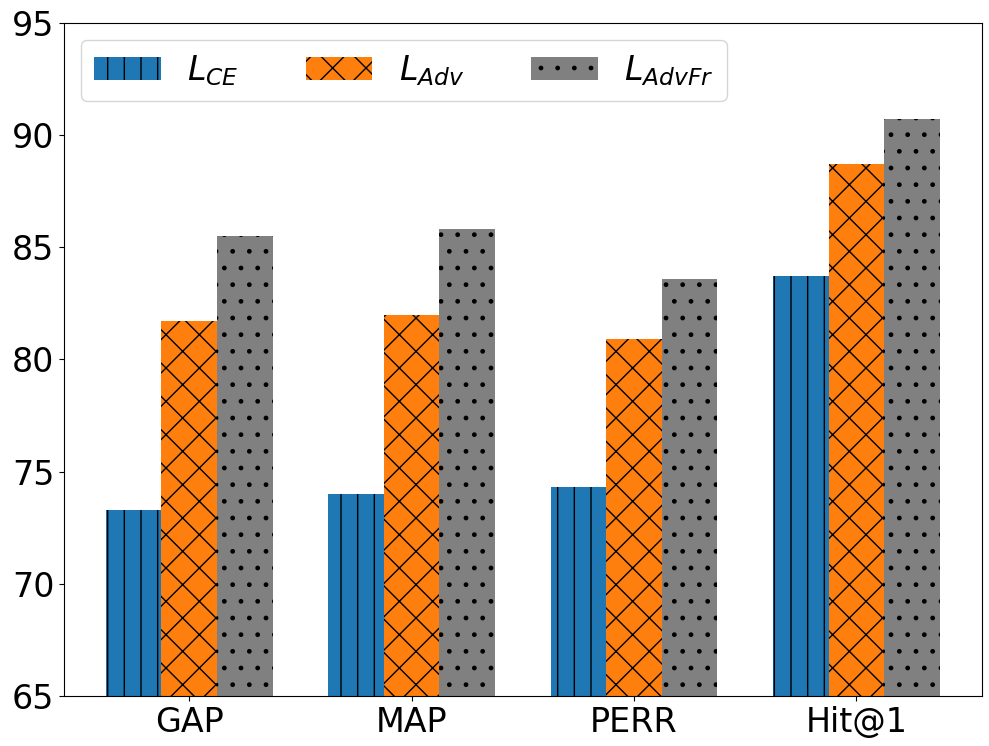}
    \caption{Performances of GMSA based models trained with $\mathcal L_{CE}$, $\mathcal L_{adv}$ and $\mathcal L_{advFr}$ when predicting the classes of adversarial examples generated using validation set samples.}
    \label{fig:adv_attack}
    \vspace{-0.5cm}
\end{figure}
\subsubsection{Qualitative analysis}
We provide some example videos and the categories predicted by baseline SA based Transformer model trained using $\mathcal L_{CE}$ and our GAT model in Table~\ref{tab:qual}. For each YouTube video, we show the ground truth and up to top 3 predictions from the baseline model and our proposed model.
Example 1 is a music video with a static poster. While our model predicts it correctly to be of `Music' category, the baseline detects it as a video game because of the image graphics. Example 2 is a video of a person dancing but the baseline model predicts it as being a `Sports' video. In example 3, the video is of people arm wrestling overlaid with a rock song playing throughout. Baseline model classifies it as `Music' whereas GAT can correctly identify that the video is about `Wrestling'. Example 4 is an interesting case. It is a video about UFOs. It is quite dark in general and only has selective frames where an UFO like object can be seen. While the baseline model incorrectly identifies it as a `Video Game', GAT can confidently predict its class correctly. 
\subsubsection{Class-wise analysis}
For each class, we computed the difference between average precision (AP) obtained using GAT and baseline model. We present the top five and bottom five classes in terms of improvement of GAT over baseline in Figure~\ref{fig:cmap}. We observe that GAT performs worse than the baseline model in only two classes out of 431. We also notice that the degradation in performance is not much. We observe that the classes which gave the least amount of increment, performances of both baseline and GAT are already quite high. It is interesting to see that some of the biggest improvements are achieved in underrepresented classes such as a specific landmark, a specific book/games genre, or science categories.
\renewcommand{\thefigure}{8}
\begin{figure}[hbtp]
\centering
    \includegraphics[width=1.0\columnwidth]{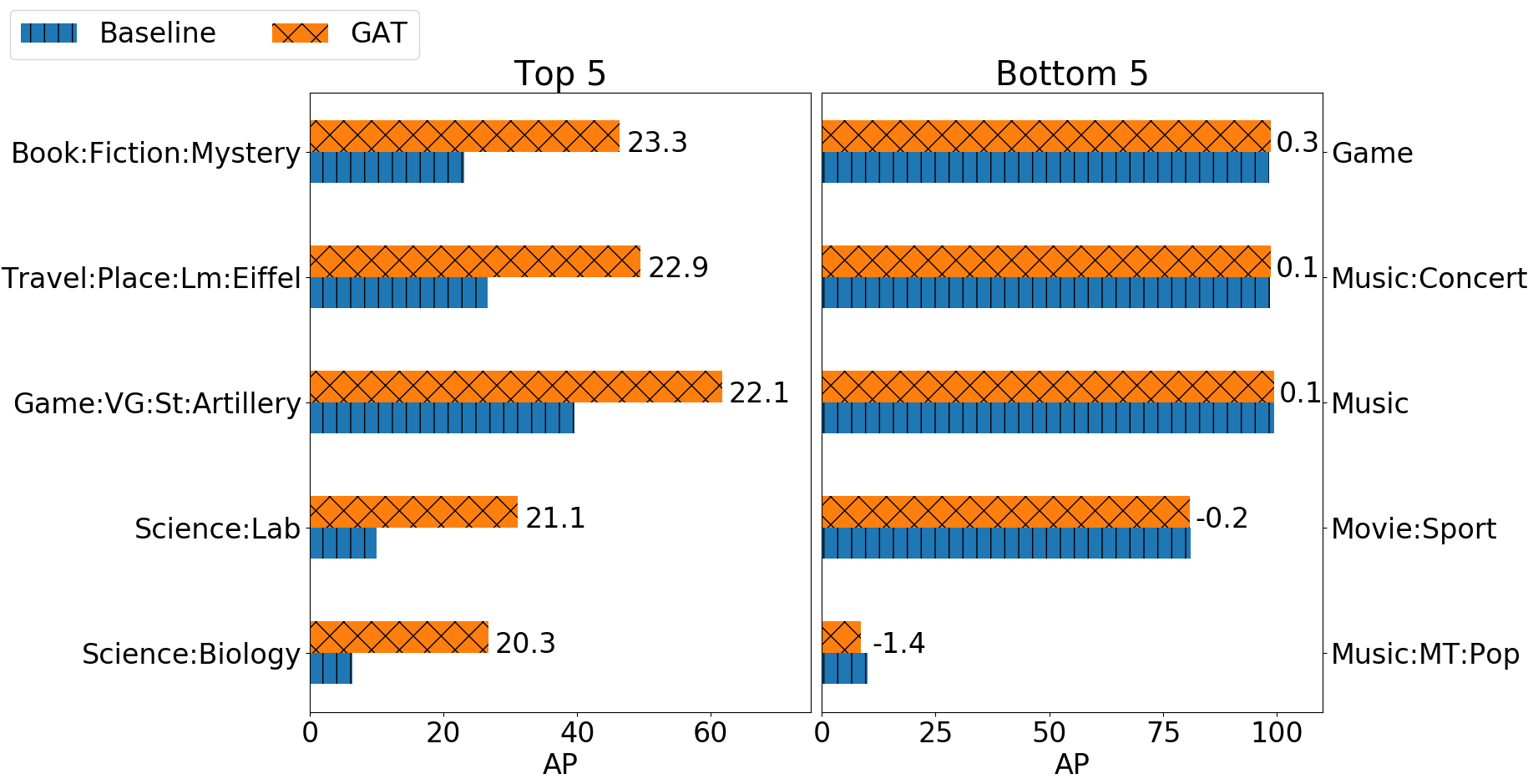}
    \caption{Average Precision (in \%) for top 5 (left) and bottom 5 classes (right). The difference between the model performances is shown beside each bar plot. Abbreviations Lm, VG, St and MT stand for Landmark, video-game, strategy and Music-Type.}
    \label{fig:cmap}
    \vspace{-0.7cm}
\end{figure}
\section{Conclusions}
In this paper, we proposed a novel approach named Gated Adversarial Transformer (GAT) for the task of video classification. We introduced a gated multi-level attention module which modifies the functionalities of a self-attention block to capture representations of a feature set based on local and global contexts. We further enhanced the performance of our model by adversarial training procedures. We introduced a regularization term in the loss function that equips the model with adversarial robustness to attention maps as well as the final output space. We performed experiments on the large scale YouTube-8M dataset and showed consistent improvements over the baseline models using our methods. We showed various ablation studies to showcase the effect of each component in the model. Further, we showed that our model not only improves performance on test data but provides significant robustness to adversarial attacks (Figure ~\ref{fig:adv_attack}). 

In the future, we would like to extend gated multi-level attention to more than two levels creating a hierarchy. Further, we would like to use video segmentation techniques to assist multi-level attention. To further improve the adversarial training aspect, we would like to include more realistic adversarial examples as well as add weighted regularization for various layers in the model.

{\small
\bibliographystyle{ieee_fullname}
\bibliography{main}
}

\end{document}